\documentclass[lettersize,journal]{IEEEtran}
\usepackage{amsmath,amsfonts}
\usepackage{algorithmic}
\usepackage{array}
\usepackage{textcomp}
\usepackage{stfloats}
\usepackage{url}
\usepackage{verbatim}

\def\BibTeX{{\rm B\kern-.05em{\sc i\kern-.025em b}\kern-.08em
    T\kern-.1667em\lower.7ex\hbox{E}\kern-.125emX}}
\usepackage{balance}                                 

\usepackage{graphics} 
\usepackage{epsfig} 
\usepackage{xcolor}
\usepackage{amssymb}
\usepackage{adjustbox}
\usepackage{booktabs} 
\usepackage{amsmath}
\usepackage[ruled,vlined]{algorithm2e}
\usepackage{hyperref}
\usepackage{xcolor}
\usepackage{graphicx}
\usepackage{amsmath,amsthm,amssymb,amsfonts, fancyhdr, color, comment, graphicx, environ}
\usepackage{xcolor}
\usepackage{mdframed}
\usepackage[shortlabels]{enumitem}
\usepackage{indentfirst}
\usepackage{hyperref}
\usepackage{subcaption}
\usepackage{hyperref}
\usepackage{wrapfig}
\usepackage{ltablex}
\usepackage{siunitx}
\newcolumntype{L}{>{\raggedright\arraybackslash}X}

\newcommand{\CP}[1]{\ignorespaces}
\newcommand{\PC}[1]{\ignorespaces}
\newcommand{\ie}{\textit{i.e.\ }}

\newcommand{\wrt}{\textit{w.r.t.\ }}

\begin{document}
\title{\LARGE \bf ORCHNet: A Robust Global Feature Aggregation approach for 3D LiDAR-based Place recognition in Orchards}

\author{T. Barros, L. Garrote, P. Conde, M.J. Coombes, C. Liu, C. Premebida, U.J. Nunes %
\thanks{T.Barros, L.Garrote, P.Conde, C.Premebida and U.J.Nunes are with the University of Coimbra, Institute of Systems and Robotics, Department of Electrical and Computer Engineering, Portugal.
E-mails:{\tt\small\{tiagobarros,~garrote,~pedro.conde,
~cpremebida,~urbano\}@isr.uc.pt}\\
M.J.Coombes and C.Liu are with the Dept. of Aeronautical and Automotive Engineering, LUCAS Lab, Loughborough University, UK.
E-mails:{\tt\small\{M.J.Coombes,~C.Liu5\}@lboro.ac.uk}
}%
}


\maketitle

\begin{abstract}
Robust and reliable place recognition and loop closure detection in agricultural environments is still an open problem. In particular,  orchards are a difficult case study due to structural similarity across the entire field. In this work, we address the place recognition problem in orchards resorting to 3D LiDAR data, which is considered a key modality for robustness. 
Hence, we propose ORCHNet, a deep-learning-based approach that maps 3D-LiDAR scans to global descriptors. Specifically, this work proposes a new global feature aggregation approach, which fuses multiple aggregation methods into a robust global descriptor. 
ORCHNet is evaluated on real-world data collected in orchards, comprising data from the summer and autumn seasons.  To assess the robustness, We compare ORCHNet with state-of-the-art aggregation approaches on data from the same season and across seasons. 
Moreover, we additionally evaluate the proposed approach as part of a localization framework, where ORCHNet is used as a loop closure detector.
The empirical results indicate that, on the place recognition task, ORCHNet outperforms the remaining approaches, and is also more robust across seasons. As for the localization, the edge cases where the path goes through the trees are solved when integrating ORCHNet as a loop detector, showing the potential applicability of the proposed approach in this task.
The code will be publicly available at:\url{https://github.com/Cybonic/ORCHNet.git} 

\end{abstract}




\section{INTRODUCTION}

Place recognition can be understood as a perception-based global localization  approach that recognizes previously visited places using visual, structural, and/or semantic cues, from which descriptors are generated. In a typical application, the current descriptor is compared with descriptors from previous input data to identify revisited locations. Lately, place recognition has been used as an efficient loop closure detector in SLAM or localization approaches\cite{cattaneo2022lcdnet,cieslewski2018data}.

In terms of research, the autonomous driving (AD) community is the most active,  using place recognition to achieve long-term localization, where data from 3D LiDARs are considered a key modality to gain robustness \cite{7747236}. Currently, 3D LiDAR-based approaches resort mostly  to deep learning (DL) methods for place modeling \cite{barros2021place}, a technique that has been widely adopted since the recent advancements in DL that let point clouds to be handled directly by the networks - an example of such a network is PoinNet\cite{Qi_2017_CVPR}.    

A less studied field in terms of place recognition and loop closure detection is field robotics, particularly, in agricultural robotics, where perception-based localization is a hard task to perform over time, due to harsh and changing conditions \cite{ou2023place}. Within the agricultural application domains, orchards are a very challenging case study because of the lack of `relevant' geometric features in the environment when compared to urban-like environments which are usually extracted from vehicles, buildings, or other static urban ``furniture". In orchards, the low density of the canopies, which may change from season to season, generates sparse scans, which leads to poor descriptive features. Moreover, the disposition of trees in parallel rows, with regular intervals, results in LiDAR scans with similar geometrical data, which makes subsequent scans, and scans from neighboring rows almost indistinguishable, a poor characteristic for approaches such as place recognition, where the goal is to learn unique and descriptive features to identify a place. 

\begin{figure}[t]
    \centering
    \includegraphics[width=\columnwidth, trim={0cm 0cm 0cm 0cm},clip]{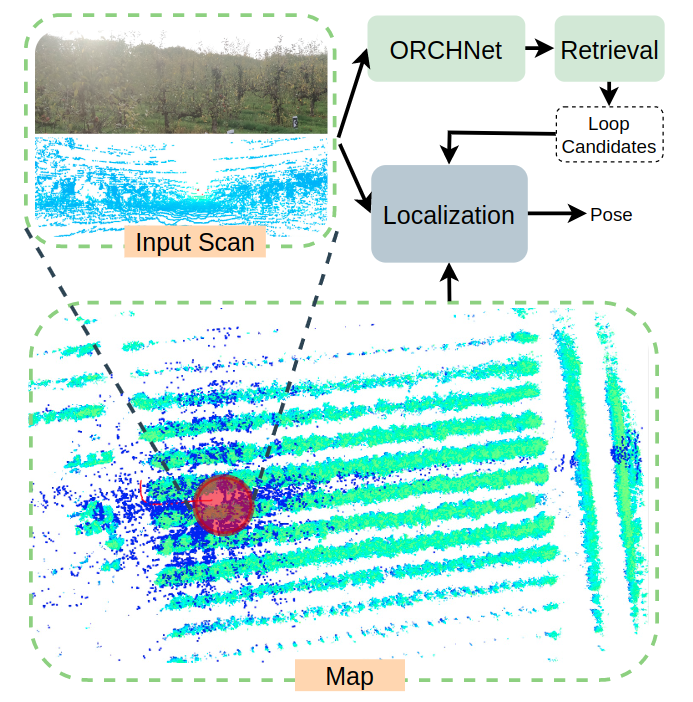}
    \caption{Representation of the proposed ORCHNet integrated in a localization framework. }
    \label{fig:localization}
\end{figure}

This work addresses this problem of place recognition and loop closure detection in orchards, proposing ORCHNet, a place modeling approach that maps 3D LiDAR scans to a descriptor space. The main contribution of ORCHNet is the global feature aggregation approach that fuses multiple aggregation approaches into a robust global descriptor. 

For this application, ORCHNet is evaluated on a self-made dataset, which was recorded in orchards in the United Kingdom, using a mobile robot equipped with sensors such as 3D LiDAR and RTK-GPS. 
The dataset comprises orchard data from the summer and autumn seasons, which allows testing the proposed approach in real changing conditions that occur in orchards. 

To evaluate the merit of ORCHNet and, in particular, the robustness across seasons in a place recognition task, two types of experiments were conducted: same season, where the training and test data are from the autumn sequence; and cross-season, where training data is from the autumn sequence and test data is from the summer sequence. ORCHNet was also evaluated as part of a localization framework (see  Fig.\,\ref{fig:localization}), where it is used to find loops, from which a qualitative assessment is presented.

The empirical results indicate that the proposed approach is more adequate for orchard than other global feature aggregation approaches. Furthermore, the results also show that the performance is not affected in cross-season experiments. As for the localization task, ORCHNet was able to solve some edge cases, where the path goes through the trees.

\noindent Succinctly, this work's key contributions are the following:
\begin{itemize}
    \item An new robust 3D LiDAR-based place recognition approach adapted for orchards/tree-containing environments.
\end{itemize}
 
\begin{figure*}[t]
    \centering
    \includegraphics[width=\textwidth, trim={0cm 0cm 0cm 0cm},clip]{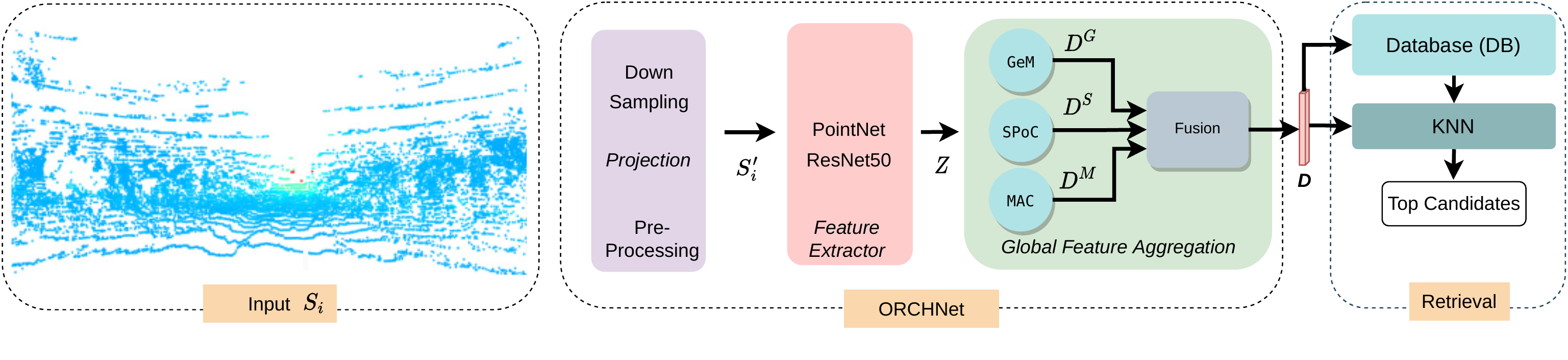}
    \caption{ORCHNet's architecture as part of a retrieval-based place recognition task. ORCHNet receives as input a point cloud $S_i$, which is down-sampled and preprocessed, returning $S'_i$. From $S'_i$, local features $Z$ are extracted, using a feature extractor. The local features are fed into the global feature aggregation module, which fuses the outputs of GeM, SPoC, and MAC into a global descriptor $D$. The global descriptor is used to query the database, which based on a similarity metric, returns the top N loop candidates.}
    \label{fig:pipeline}
\end{figure*}


\section{RELATED WORK}
Place recognition has been the subject of much research over the last decade, where 3D LiDAR approaches have been a very active topic \cite{barros2021place}.  3D-LiDAR sensory data has been considered a key modality to achieve robustness in place recognition due to being invariant to challenging visual changing conditions, which may arise during long-term operation, especially when revisiting places in different seasons and illumination conditions. 

3D-LiDAR place recognition approaches have been a natural response to overcome the challenges that visual-based approaches face. Within the 3D-LiDAR-based approaches, Deep Learning (DL) methods are the most common for place modeling. The point cloud-based DL approaches can be split into two major subfields: those that extract features directly from point clouds, such as  PointNetVLAD \cite{angelina2018pointnetvlad} or Minkloc3d \cite{komorowski2021minkloc3d}, and those that extract features indirectly from point clouds, projecting the point clouds to a proxy representation, such as voxels \cite{siva2020voxel}, polar coordinates \cite{li2022rinet} or depth range images \cite{barros2022attdlnet}. In both approaches, these inputs are fed to a feature extraction module, with the goal of extracting local features, which are, then, aggregated into a global descriptor. 

Despite the natural robustness towards appearance-changing conditions, 3D-LiDAR-based approaches have still some limitations, such as generating a global descriptor that is invariant to rotation, which is essential when revisiting a place from the opposite direction. Thus, generating a global descriptor that is invariant to rotations but simultaneously is descriptive enough to identify a place, is a fine balance that the models have to learn from the data. In this process, the feature aggregation module is of a great importance, given that it is in the aggregation step that the local features are converted into a global descriptor, where the networks define which features are the most important from a global perspective. In this regard, several aggregation approaches have been suggested for 3D-LiDAR data, where NetVLAD \cite{angelina2018pointnetvlad,arandjelovic2016netvlad} has been one of the most popular. NetVLAD splits the feature space into clusters to generate a global descriptor. Other approaches comprise attention \cite{qi2017pointnet}, generalized-mean pooling (GeM)\,\cite{komorowski2021minkloc3d},  global max pooling (MAC)\,\cite{tolias2015particular}, average-pooling (SPoC)\,\cite{7410507}.

This work leverages the GeM, SpoC, and MAC aggregation approaches, which are efficient and have shown to be adequate for point clouds, fusing their respective outputs into a global descriptor.


\section{Proposed Approach}

\begin{figure*}[t]
    \centering
    \includegraphics[width=\textwidth, trim={0cm 0cm 0cm 0cm},clip]{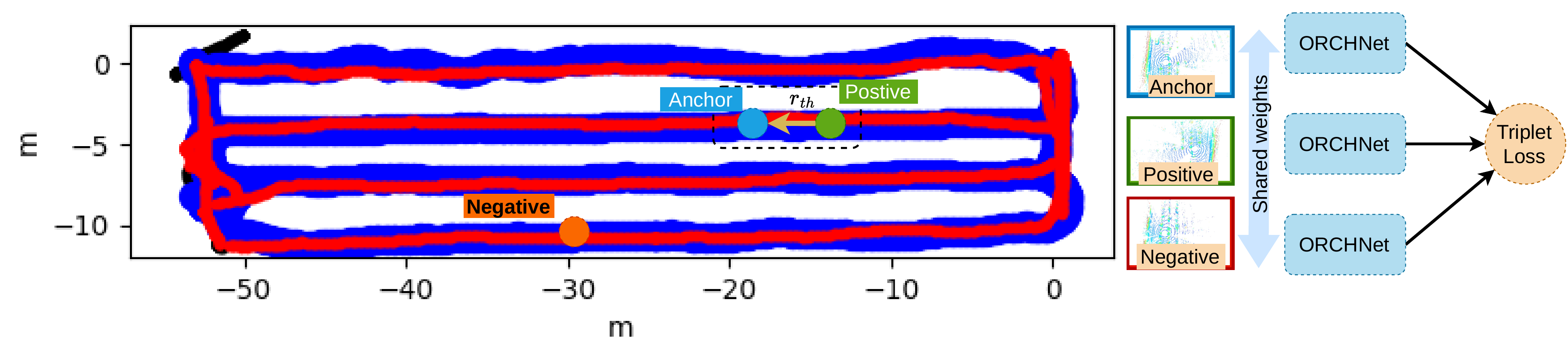}
    \caption{Training approach using the LazyTriplet loss. The anchor-positive pair has to be from the same physical (\ie close within the same line), and the anchor-negative pair has to be from distinct places. The LazyTriplet loss is computed by measuring the similarity distance between the anchor-positive and the anchor-negative pair.}
    \label{fig:train_pipeline}
\end{figure*}

This section details the proposed ORCHNet in a retrieval-based place recognition framework. An overview of ORCHNet and the framework is illustrated in Figure \ref{fig:pipeline}. The place recognition framework has two main modules: the ORCHNet, which models places by mapping point clouds to a descriptor space; and the Retrieval module. The modeling is achieved through the following main steps: an input point cloud $S_i$ is preprocessed and fed to a feature extractor, which returns local features $Z$. The local features are aggregated into a global descriptor $D$, using the proposed global feature aggregation module, which is the main contribution of this work.\\
\indent Observation: The sets defined in the following subsections are ordered sets (\ie, there is a relation of order between the elements of the set) and this order is defined by the indices of the elements.

\subsection{Pre-processing \& Feature Extractor}
 The aim of the pre-processing and feature extracting modules is to extract adequate features from the scans, by mapping them to an intermediate feature space.
Hence, given a scan  $S_i \in \mathbb{R}^{n_i \times 3}$ with $i \in [1,...,N]$ where $N$ is the number of scans in the sequence and $n_i$ the number of points in the $i^{th}$ point cloud, the feature extractor maps $S_i$ to a fixed feature vector $Z_i \in \mathbb{R}^{n \times F}$ with $F\gg 3$. In order to compare different feature extraction approaches, this work resorts to an extractor that handles point clouds directly (PoinNet), and another that extracts feature from an image-like proxy representation (ResNet50). ResNet50 was initially used on images, but has also been used on other modalities with an appropriate projection to an image-like representation. Both networks were slightly changed to fit this application.

\subsection{Global Feature Aggregation}

The proposed global feature aggregation approach takes advantage of three different aggregation methods: MAC\cite{tolias2015particular}, GeM\cite{radenovic2018fine,komorowski2021minkloc3d} and SPoC\cite{7410507}, fusing their outputs into a global descriptor  using a trainable weighted sum. The proposed approach is illustrated in Fig.\ref{fig:pipeline}. It receives a feature vector $Z \in \mathbb{R}^{N \times F}$ and outputs a descriptor $D = [d_k] \in \mathbb{R}^{K}$, being a function $\mu:\mathbb{R}^{N \times F} \rightarrow  \mathbb{R}^K$.

\subsubsection{MAC} 
The MAC module is a global max pooling operator, which computes from input features  $Z=[z_{i,j}] \in \mathbb{R}^{N\times F}$, the maximum activation value of the $i^{th}$ dimension:
\begin{equation}
    z^M_i = \max_{j}\, z_{i,j}, \ \forall i \in [1,...,N]
\end{equation}
\noindent with $Z^M = [z^M_i] \in \mathbb{R}^N$. Then, $Z^M$ is fed to a full-connected function, $h^M : \mathbb{R}^N \rightarrow \mathbb{R}^K$, which outputs the MAC's output descriptor $D^M =[d^M_k]$ with $K$ dimensions.

\subsubsection{SPoC} 
The SPoC module is an average pooling operator, which computes, from the input features $Z=[z_{i,j}] \in \mathbb{R}^{N \times F}$, the average activation value of the $i^{th}$ dimension :
\begin{equation}
    z^S_i = \frac{1}{F} \sum_{j=1}^{F} z_{i,j}, \ \forall i \in [1,...,N]
\end{equation}
\noindent with $Z^S = [z^S_i] \in \mathbb{R}^N$. Then, $Z^S$ is fed to a full-connected function $h^S: \mathbb{R}^N \rightarrow \mathbb{R}^K$, which outputs the SPoC's output descriptor $D^S=[d^S_k]$ with $K$ dimensions.

\subsubsection{GeM} 
The GeM module is a generalized-mean pooling operator that is applied to the input feature vector $Z=[z_{i,j}] \in \mathbb{R}^{N \times F}$: 
\begin{equation}
    z^G_i = \left( \frac{1}{F} \sum_{j=1}^{F} z_{i,j}^p \right)^{1/p}, \ \forall i \in [1,...,N]
\end{equation}

\noindent where $p \in \mathbb{R}$ is a trainable parameter and with $Z^G = [z^G_i] \in \mathbb{R}^N$. Then, $Z^G$ is fed to a full-connected function $h^G:  \mathbb{R}^N \rightarrow  \mathbb{R}^{K}$, which outputs the GeM's output descriptor  $D^G = [d ^G_k]$  with $K$ dimensions.

\subsubsection{Fusion}
The fusion module merges the three descriptors ($D^M$,$D^S$ and $D^G$) into a global descriptor $D$,  using  $A = [a_1, a_2, a_3] \in \mathbb{R}^3$, which are trainable parameters that learn the best activation combinations. First, the \CP{three} descriptors are concatenated into $D'=[D^M,D^S,D^G]^T = [d'_{i,k}]\in  \mathbb{R}^{3 \times K}$, then a weighted sum is computed:
\begin{equation}
    d_k = \sum_{i=1}^3 a_i d'_{i,k}, \ \forall k \in [1,...,K]
\end{equation}

\noindent with $D = [d_k]  \in  \mathbb{R}^K$, which is the $K$-dimensional output descriptor of the proposed ORCHNet.

\begin{figure*}[ht]
    \centering
    \includegraphics[width=\textwidth,trim=0 0 0cm 0, clip]{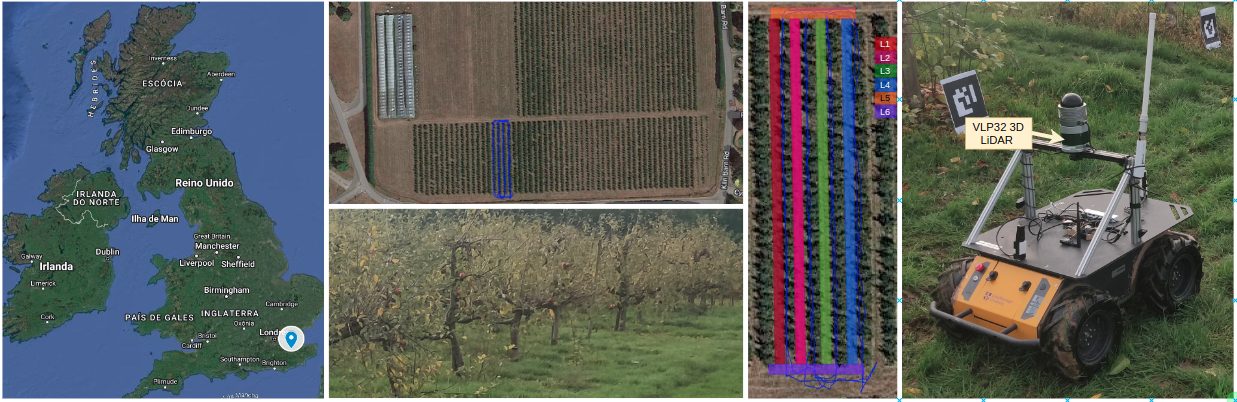}
    \caption{Illustration of the orchards and the robot platform used to collect data. The orchards are located in the southern part of the United Kingdom. The orchards were split in rows, here the autumn sequence is illustrated, which was split into 6 rows. }
    \label{fig:robot}
\end{figure*}

\subsection{Network Training} \label{sec:loss}

In this  work, the global descriptors are trained using the LazyTriplet loss as proposed in \cite{angelina2018pointnetvlad}. The LazyTriplet loss forces the network to learn a descriptor space where descriptors from the same physical place (positive samples \wrt the anchor) are close, while descriptors originated from different places (negative samples \wrt the anchor) are apart.

Hence, given a collection of scans $\mathcal{S} = \{S_i| S_i \in \mathbb{R}^{n_i\times 3} \}$ and the corresponding poses $\mathcal{P} = \{p_i | p_i \in \mathbb{R}^{3}\}$ with $i = [1,...,N]$ where $N$ is the number of samples in the sequence, and $n_i$ is the number of points in the $i^{th}$ scan. Let us consider a set $\Gamma = \{(S_i,p_i)|S_i \in \mathcal{S}, p_i \in \mathcal{P}\} \subseteq \mathcal{S} \times \mathcal{P}$, which defines that for each scan $S_i$ exists a corresponding pose $p_i$. 

Let us define a set $\Gamma^A \subseteq \Gamma$ where each element  $(S^A_j, p^A_ j) \in \Gamma^A$ is an anchor (\ie for which exist a corresponding positive loop) with $j \in [1,...,J]$ and $J<N$ is the number of anchors in the sequence $\mathcal{S}$. 

Let us also define (for some parameter $\gamma)$ a set of positives $\Gamma^{P_{j}}_\gamma  \subseteq \Gamma $ ($ \forall j \in [1,...,J]$) \wrt the $j^{th}$ anchor $(S^A_j,p^A_j)$. First, we define the function $\omega: \Gamma \rightarrow \mathbb{N}$, that maps each pair of $\Gamma$ to the its row number. Now we can formally define
\begin{align}
\label{set:positives}
    \Gamma^{P_{j}}_\gamma = \Big{\{}(S^{P_j}_l,p^{P_j}_l)\,| & \  \|p^{P_j}_l-p_j^A\|_2 < r_{th} \nonumber \\ 
    & \wedge \ \omega(S^{P_j}_l,p^{P_j}_l) = \omega(S^A_l,p^A_l)\\
    & \wedge l \notin [j-\gamma, j]  
    \qquad \qquad \qquad \Big{\}} \nonumber
\end{align}
\noindent where each element  $(S^{P_j}_l,p^{P_j}_l)  \in \Gamma^{P_{j}}_\gamma$, is a positive \wrt the $j^{th}$ anchor. In summary, the positives defined by $\Gamma^{P_{j}}_\gamma$ respect three different conditions, described (by this order) in equation \eqref{set:positives}: belong to a neighborhood of the $j^{th}$ anchor, defined by a radius $r_{th}$; belong to the same row as the $j^{th}$ anchor; do not belong to set of the $\gamma$ previous elements of $\Gamma$, \wrt to the $j^{th}$ anchor.

And, finally, let us also define a set of negatives $\Gamma^{N_{j}}  \subseteq \Gamma $ \wrt the $j^{th}$ anchor $(S^A_j,p^A_j)$. The negatives are samples that are outside the region of interest defined by $r_{th}$\,, \ie
\begin{equation}
    \Gamma^{N_{j}} = \{(S^{N_j}_m,p^{N_j}_m)\, |\  \|p^{N_j}_m-p_j^A\|_2 \geq r_{th}\} 
\end{equation}
\noindent where each element  $(S^{N_A}_m,p^{N_A}_m) \in \Gamma^{N_{j}}$ is a negative \wrt the $j^{th}$ anchor. In practice, the set of negatives is generated by taking a random sample of the original $\Gamma^{N_{j}}$.

Now to compute the loss for a given anchor $(S^A_j, p^A_j)$, the LazyTriplet loss is computed by the following expression:

\begin{equation}
    \mathcal{L}_T= \text{max}(\|d^A_j-d^P_{c_1}\|_2-\|d^A_j-d^N_{c_2}\|_2+m,0)
    \label{eq:lt}
\end{equation}

\noindent where $d^A_j$ is the descriptor of the $j^{th}$ anchor scan $S^A_j$;  $d^P_c$ is the descriptor of the closest  positive (in the physical space) $(S^P_{c_1},p^P_{c_1}) \in \Gamma^P$  \wrt $(S^A_j,p^A_j)$, \ie ${c_1} = \text{arg\,min}_l\  \| p^P_l - p^A_j \|_2$; 
and $d^N_c$ is the descriptor of the closest negative (in the descriptor space) $(S^P_{c_2},p^P_{c_2}) \in \Gamma^P$ \wrt  $(S^A_j,p^A_j)$, \ie $c_2 = \text{arg\,min}_m\  \| d^P_m - d^A_j \|_2$. Finally, $m$ is a margin value. This process is illustrated in Fig. \ref{fig:train_pipeline}.

\section{EXPERIMENTAL EVALUATION} \label{sec:experiments}

The aim of this work is to evaluate the robustness of place recognition in orchards, namely one of the goals is to assess the robustness of the proposed approach over different seasons. Hence, to support the proposed contributions, this section presents a self-made orchard dataset, the experiments and evaluation protocols, the implementation details, and finally, the results obtained on a retrieval task, and as part of a localization framework are discussed. 

\subsection{Orchard Dataset} \label{sec:gt}
The dataset comprises two sequences, which were recorded in the same Bramley apple orchard located in Kent in the United Kingdom, but in different seasons (\ie summer and autumn), using a Clearpath Husky mobile robot equipped with a Velodyne VLP32 3D LiDAR (10Hz) and a ZED-F9P RTK-GPS (5Hz).  These two sequences capture the trees in different flowering states, which introduces an additional challenge, due to the changing conditions of the trees and the environment. Both were conducted on bright dry days with low wind speeds. Figure \ref{fig:robot} illustrates this robot with the sensors during a recording session. For the purpose of this work, a loop exists when scans from different revisits are within a range of 10m: an anchor-positive pair exists when two scans, from different revisits, are within a range of 10m.

\begin{figure}[t]
    \centering
    \includegraphics[width=\columnwidth, trim={0cm 0cm 0cm 0cm},clip]{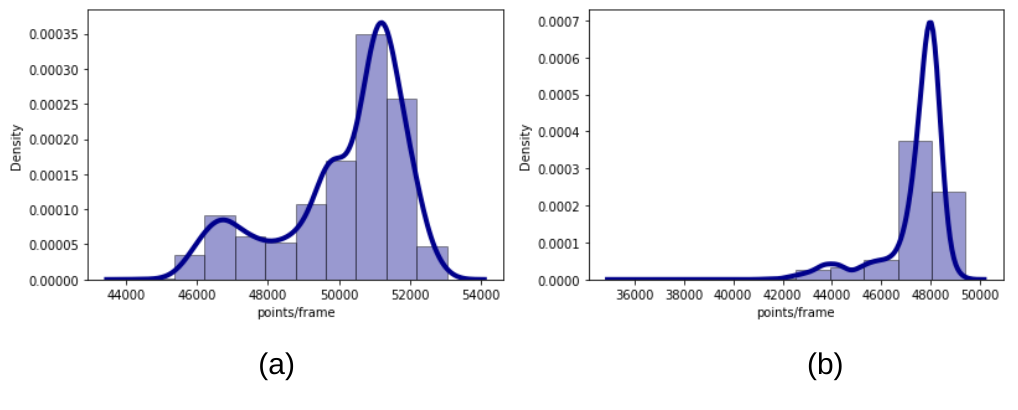}
    \caption{Illustration of the points per frame distributions of a) sequence summer and b) sequence autumn.}
    \label{fig:points_per_frames}
\end{figure}

\begin{figure}[t]
    \centering
    \includegraphics[width=\columnwidth, trim={0cm 0cm 0cm 0cm},clip]{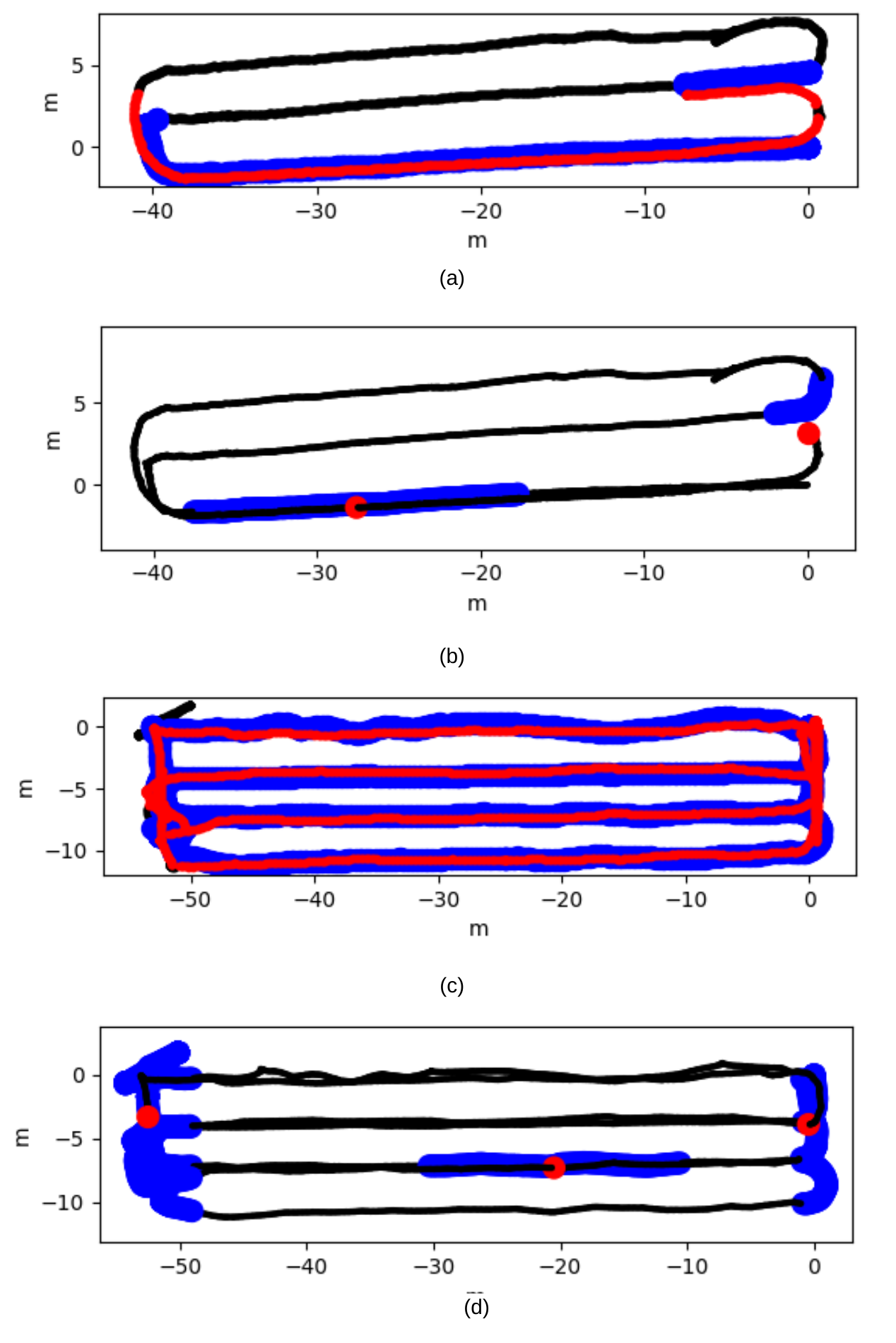}
    \caption{Illustration of the ground-truth paths. Figures a) and c) outline the loops of the summer and autumn sequences respectively, where anchors are highlighted in red, positives in blue, and the remaining points in black. Figures b) and d) outline the positives within 10\,m for a given anchor of the summer and autumn sequences, respectively, where the anchor is highlighted in red and the positives in blue.}
    \label{fig:ground-truth}
\end{figure}

In both sequences, the robot traveled in the orchard lines defining paths containing revisited segments from the same and opposite directions. The summer sequence was recorded in July and comprises 3244 scans with synchronized poses. From these scans, 954 are anchors. While the summer sequence has segments that are revisited only once, the autumn sequence has segments with several revisits. The autumn sequence was recorded in November and has 3674 scans with synchronized poses, from which 2311 are anchors. Figure\,\ref{fig:points_per_frames} shows the point distributions per scan in each sequence, while the ground-truth loops are illustrated in Fig.\,\ref{fig:ground-truth}.a) and Fig.\,\ref{fig:ground-truth}.c).  Table\,\ref{tab:sequence} summarizes the information regarding the scans and existing loops in each sequence. Figure\,\ref{fig:ground-truth}.b) and Fig.\ref{fig:ground-truth}.d) illustrates for a given anchor, the respective positives within the range of 10\,m for the summer and autumn sequence, respectively. 

\begin{table}[t]
  \centering
  \caption{\CP{Dataset and Loop information} Number of frames and loops in the dataset. The number loops correspond to a $r_{th}\le 10m$.}
  {\renewcommand{\arraystretch}{1.5}
	\begin{adjustbox}{max width=\columnwidth}
        \begin{tabular}{p{0.3\columnwidth}p{0.2\columnwidth}p{0.2\columnwidth}p{0.2\columnwidth}}
        \noalign{\hrule height 1pt}\hline	
        Sequence & Length [frames]   & Loops [frames] & Points/Frame \\
        \midrule
		\midrule
        Autumn  & 3674  &  2311 & 47k$\pm$1k \\
        Summer  & 3244  &  954  & 50k$\pm$2k  \\
       \noalign{\hrule height 1pt}\hline	
        \end{tabular}%
    \end{adjustbox}
    }
  \label{tab:sequence}%
\end{table}%

\subsection{Evaluation of Place Recognition}
 
The evaluation of the proposed approach is conducted using the standard retrieval metric which is Recall, defined as follows:
 \begin{equation}
     {\renewcommand{\arraystretch}{1.5}
     \text{Recall} = \frac{\text{TP (\# Retrieved Loops)}}{\text{TP + FN (\# Loops)}},
     }
    \label{eq:recall}
\end{equation}
namely reporting the recall of the most similar retrieved candidate (\ie, recall@1) and the recall of the 1\% most similar candidates (\ie, recall@1\%). 

Hence, to evaluate the performance of the proposed approach, the following retrieval protocol was implemented. Let us first consider the set of scans and corresponding poses $\Gamma = \{(S_i,p_i)|S_i \in \mathcal{S}, p_i \in \mathcal{P}\}$. For each element $(S_i,p_i)\in \Gamma$ we will denote by $D_i$ the descriptor of  $(S_i,p_i)$ generated by the model, defining the set of descriptors $\mathcal{D} =\{D_i| D_i \in \mathbb{R}^{K}\}$, with $K$ dimensions and $|D|=|\Gamma|$.  Let us also define  a database $\Theta_t = \{D^{\Theta_t}_m | D^{\Theta_t}_m \in  \mathbb{R}^{K} \} \subseteq \mathcal{D}$  with $m \in [1,...,M_t]$, where $M_t$ is the number of descriptors in the database at iteration $t$. \\
\indent Moreover, we will also consider the previously defined set of anchor pairs $\Gamma^A$,  to define $\mathcal{D}^A =\{D^A_j \in D | (S^A_j,p^A_j) \in \Gamma^A\} \subseteq \mathcal{D}$, \ie the set of descriptors associated to the anchors $(S^A_j,p^A_j) \in \Gamma^A$; let us note that $|D^A|=|\Gamma^A|$. \PC{$\mathcal{D}$ that are anchors, and $j \in [1,...,J]$, where $J$ is the number of anchors in the sequence.}

Given an anchor's descriptor $D^A_j \in \mathcal{D}^A$ and the database $\Theta_t$, the set of $N$ top candidates $\mathcal{D}^j_{N,t}$  \wrt the $j^{th}$ anchor are retrieved using a K-nearest neighbor(KNN) method from $\Theta_t$ and defined as follows:  $\mathcal{D}^j_{N,t} =\{d_n \in \Theta_t | \| d_n - d^A_j \|_2 \leq \| d_{n+1} - d^A_j \|_2 \ \wedge \ n \in [1,...,N]   \}  $\PC{ with $n \in [1,...,N]$ where $N$} \text{ } (with $|\mathcal{D}^j_{N,t}| = N$) \ie the set of the $N$ descriptors of $\Theta_t$ that are closer (in the descriptor space) to the descriptor of the $j^{th}$ anchor.\\
\indent Hence, to compute recall@N of a given model (with N representing the top N candidates), given a set of top candidate descriptors $\mathcal{D}^j_{N,t}$, we start by defining $\Gamma^j_{N,t}=\{(S_n,p_n) \in \Gamma| d_n \in \mathcal{D}^j_{N,t} \}$ (with $|\Gamma^j_{N,t}|=|\mathcal{D}^j_{N,t}|=N$), \ie the set of pairs (scans and poses) associated with the set of top candidate descriptors. Then, the number of true positives (in the Equation \eqref{eq:recall}), \wrt the $j^{th}$ anchor, is defined by $|\Gamma^j_{N,t} \cap \Gamma^{P_{j}}_\gamma|$, \ie the number of common elements in the set of top candidates ($\Gamma^j_{N,t}$) and set of positive samples ($\Gamma^{P_{j}}_\gamma$), previously defined.

The proposed approaches are compared with state-of-the-art  global feature aggregation approaches such as NetVLAD \cite{angelina2018pointnetvlad}, GeM\cite{komorowski2021minkloc3d} MAC \cite{komorowski2021minkloc3d}, and SPoC\cite{7410507}. Thus, in order to establish a fair comparison among the various methods, and assess the merit of each approach, all global feature aggregation methods are evaluated on the same feature extractor (\ie backbone)  network. 
As for  feature extractor, two approaches are used in this work: PointNet\cite{Qi_2017_CVPR} and ResNet50\cite{he2016deep}. These two networks process the point clouds differently, while PointNet is a neuronal network that learns features directly from the point cloud, ResNet50, on the other hand, is a CNN-based network, which was initially proposed for images, requiring thus to project point clouds to a proxy representation such as spherical or Bird's-eye View (BEV). In this work, in ResNet50-based experiments, the point clouds are projected to the BEV representation. Comparing the two feature extractors allows a deeper assessment of the proposed approach when fed with distinct input representations.

\subsection{Implementation and Training Details}

All models were trained for 100 epochs on a NVIDIA GeForce RTX 3090 GPU, using the closest positive and 20 negatives. The margin value  $m$ was set to 0.5, and the model parameters were optimized using the AdamW optimizer with a learning rate ($Lr$) of 0.0001 and a weight decay ($Wd$) of 0.0005.  Moreover, all proposed experiments were conducted on Python 3.8, using the PyTorch with CUDA 11.6 for the networks.

In all experiments, if not otherwise specified, the point clouds were cropped along the x-axis and y-axis at $\pm15m$ respectively, and then  down sampled to 20k points for PointNet and to 512 points for ResNet50, which are randomly selected. Additionally, at each training step, the point clouds are augmented by applying a random rigid body transformation with a maximum rotation of $\pm180^\circ$.

 Furthermore, as suggested in \cite{angelina2018pointnetvlad}, the number of clusters of NetVLAD is set to 64. The trainable parameters $A=[a_1,a_2,a_3]^T$ are initialized based on a normal distribution with $\mu = 0$ and $\sigma = 0.1$. Moreover, the configuration of each network was adjusted in order to return the best performance. For instance, ResNet50 presented the best performance with a global descriptor size of 512 dimensions, while PointNet required 2048 dimensions.  

\subsubsection{PointNet-based}
PointNet is a neuronal network that learns features directly from the point clouds. In this work, PointNet is cropped before the max pooling layer, as proposed in \cite{angelina2018pointnetvlad}. Despite learning from point clouds directly, PointNet, nevertheless, needs a fixed size input, hence, instead of using the raw scan directly as input, $S_i$ is down sampled to $S_i^{\prime}$.

PointNet receives as input a tensor $S'\in  \mathbb{R}^{b\times 1 \times n' \times F}$, where $b$ is the batch size, $n'$ the number of points (already down sampled), and F the feature dimensions. In this work, the batch is set to $20$ for testing and $1$ for training, the number of points is set to $20k$, and $F$ is set to $3$. The first layer of the network maps this 3-dimensional tensor to 1024 dimensions, which are down-scaled to 512 dimensions at the PointNet's output (\ie $Z \in \mathbb{R}^{b\times 20k \times 512}$).  The global feature aggregation module receives this $Z$ tensor and output a descriptor $D \in \mathbb{R} ^{b \times 2048}$.
 
\subsubsection{ResNet50-based}

ResNet50 is a CNN-based network, which extracts feature from image-like representations. In this work, when using ResNet50 as feature extractor, point clouds are projected to a BEV representation, which is obtained by discretizing the point clouds into a 2D grid, where the cells are populated with the height  ($z$ coordinate), the point density, and the intensity that fall into the corresponding cells. 
Hence, the process of extracting features from point clouds using ResNet50 is the following: first the scan $S_i \in \mathbb{R}^{n \times 3}$ is down sampled to $S'_i \in \mathbb{R}^{500 \times 3}$, which are projected to an image $B_i = [R_h,R_d,R_i] \in \mathbb{R}^{b \times 3 \times h \times w}$, which is the concatenation of the  height ($z$ coordinate) $R_h \in \mathbb{R}_h^{h\times w}$, the point density $R_d \in \mathbb{R}^{h\times w}$ and the intensity $R_i \in \mathbb{R}^{h \times w}$ of the scan's projection into the BEV representation, where $b$ is the batch size, and $h$ and $w$ are the image's height and width, respectively. In this work, $h$ and $w$ are set to 256 each, and the batch is set to 20 during testing, and to 1 during training.  
ResNet50 uses $B_i$ as input and outputs a tensor $Z' \in \mathbb{R}^{b \times 2048 \times 16 \times 16}$, which is reshaped to $Z \in   \mathbb{R}^{b \times 2048 \times 256}$. Then, $Z$ is fed to the global feature aggregation module, which outputs a global descriptor $D \in \mathbb{R} ^{b \times 512}$.

\begin{table}[t]
  \centering
  \caption{\CP{Dataset and Loop information} Number of frames and loops in the dataset.}
  {\renewcommand{\arraystretch}{1.5}
	\begin{adjustbox}{max width=\columnwidth}
        \begin{tabular}{p{0.2\columnwidth}p{0.3\columnwidth}p{0.3\columnwidth}p{0.2\columnwidth}}
\noalign{\hrule height 1pt}\hline	        Experiments  & Sequences  & Training set & Test set \\
        \midrule
		\midrule
        Same season  & Autumn  & 60$\%$ (1386)  &  40$\%$ (925)    \\
        Cross season & Autumn/Summer  &  2311 &  954 \\
\noalign{\hrule height 1pt}\hline	        \end{tabular}%
    \end{adjustbox}
    }
  \label{tab:experiments}%
\end{table}%

\subsection{Retrieval Performance} 
\begin{table}[t] 
\caption{Empirical results obtained on the same-season experiment, where the autumn sequence was split into 60\% for the training set and 40\% for the test set.} \label{tab:sameseason}
{\renewcommand{\arraystretch}{1.5}
\begin{adjustbox}{max width=\columnwidth}
\setlength{\tabcolsep}{10pt} 
\begin{tabular*}{\columnwidth}{l|cc|cc}
\noalign{\hrule height 1pt}\hline	

                   & \multicolumn{2}{c|}{Recall@1} & \multicolumn{2}{c}{Recall@1\%}                                     \\
                   & PointNet & ResNet & PointNet & ResNet   \\ \midrule \midrule
VLAD  & 0.11 & 0.41 & 0.68 & 0.92 \\
SPoC  & 0.50 & 0.44 & 0.85 & 0.93 \\
GeM   & 0.49 & 0.41 & 0.84 & 0.93 \\
MAC   & 0.49 & 0.44 & 0.84& 0.92  \\  \hline
ORCHNet (our)  & \textbf{0.52} &\textbf{0.45} & \textbf{0.92}& \textbf{0.94  }\\ \noalign{\hrule height 1pt}\hline	

\end{tabular*}
\end{adjustbox}
}
\end{table}

\begin{table}[t]
\caption{Empirical results obtained on the cross-season experiment, where the autumn sequence was used for training and the summer sequence for testing.} \label{tab:freq}
{\renewcommand{\arraystretch}{1.5}
\begin{adjustbox}{max width=\columnwidth}
\setlength{\tabcolsep}{10pt} 
\begin{tabular*}{\columnwidth}{l|cc|cc}
\noalign{\hrule height 1pt}\hline	                   & \multicolumn{2}{c|}{Recall@1} & \multicolumn{2}{c}{Recall@1\%}                                     \\
                   & PointNet & ResNet & PointNet & ResNet   \\ \midrule \midrule
VLAD  & 0.061 & 0.35 & 0.58 & 0.92 \\
SPoC  & 0.43 & 0.41 & 0.76 & 0.93 \\
GeM   & 0.43 & 0.38 & 0.79 & 0.93 \\
MAC   & 0.40 & 0.40 & 0.79 & \textbf{0.96 } \\  \hline
ORCHNet (our)  & \textbf{0.50}& \textbf{0.42}  & \textbf{0.85}& 0.95 \\  \noalign{\hrule height 1pt}\hline	
\end{tabular*}
\end{adjustbox}
}
\label{tab:crossseason}
\end{table}
This section presents and discusses the empirical results obtained with ORCHNet. 
Two main experiments were conducted: same season and cross season. As for the experiments on data from the same season, the autumn sequence was split into $60\%$ for training and $40\%$ for testing. As for the cross-season experiments, the models were trained on the autumn sequence, while tested in the summer sequence. Table \ref{tab:experiments} summarizes the split in terms of training and testing samples for each experiment.  

The results obtained on the same season, which are presented in Table \ref{tab:sameseason}, indicate that all methods, ORCHNet included, have a poor performance when retrieving only one candidate (\ie recall@1). The reason behind this low performance can be attributed to the low geometrical differences in orchards, which leads to the networks' incapacity to extract descriptive features in order to distinguish the scans from the various lines. On the other hand, when retrieving the top $1\%$ (of the elements in the database) as loop candidates, which in this experiment were 31 candidates, the performance grows considerably to almost perfect recall. This means that, despite the ambiguity in the top 1 candidate, when retrieving $1\%$, all networks are able to achieve reasonable performance, with ORCHNet outperforming the remaining methods.

 The results obtained on the cross-season experiment, which are presented in Table  \ref{tab:crossseason},  indicate that, despite all models loose in performance,  which was expected, PointNet-based approaches still perform better at top@1 retrieving, while ResNet50-baed approaches have higher performance at top 1\%. The results also indicate that ORCHNet is able to maintain performance across seasons, showing thus that it is robust to seasonal changes.

\begin{figure}
    \centering
    \includegraphics[width=\columnwidth, trim={0cm 0cm 0cm 0cm},clip]{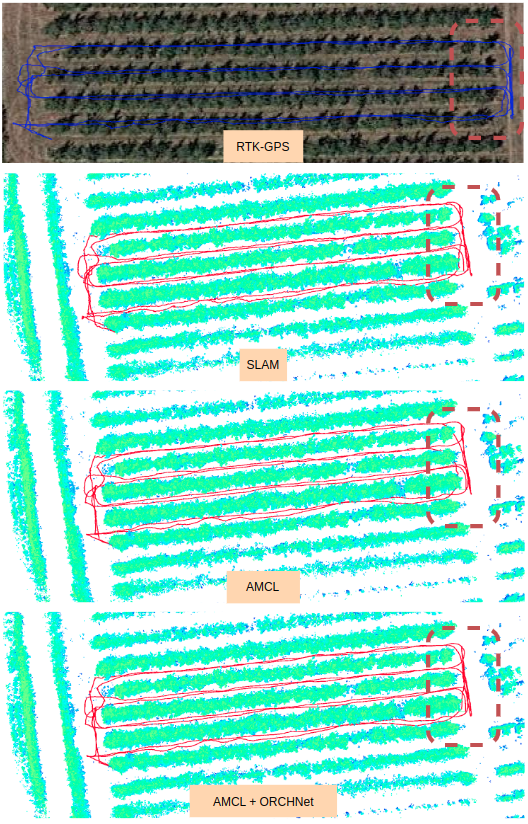}
    \caption{Autumn sequence and localization framework results. From top to bottom: RTK-GPS data, HMaps-based SLAM, AMCL and AMCL with place proposals provided by the ORCHNet.}
    \label{fig:maps}
\end{figure}

\subsection{Evaluation on Localization Framework} 
ORCHNet was additionally evaluated as part of an Adaptive Monte Carlo Localization (AMCL) framework, for which qualitative results from the autumn sequence are reported. The evaluation consists in comparing the generated paths from the AMCL (only), AMCL-ORCHANet, SLAM, and RTK-GPS.
First, an HMaps-based SLAM approach as proposed in \cite{garrote2018hmaps}  was used to create an environment representation of the autumn sequence (see Fig. \ref{fig:maps}).
Then, the AMCL approach is adapted to take into account a set of loop proposals provided by the ORCHNet model. The adaptation was made in the AMCL's resampling stage, where a stratified approach as proposed in \cite{garrote2019amclstrat} was used to select (i) if a given particle is sampled from the current set of particles or (ii) if it has been sampled from the set of loop candidates.
The sampling of loop proposals is conditioned on the similarity score of each loop candidate, meaning that similar place proposals have a higher sampling probability. This is achieved with a multinomial resampling approach. Additionally, a scan-matching-based LiDAR odometry approach is used to predict the particles' state. As for the weight update, it is important to note that only a small number of points (1 in 50) is used to compute each particle's state.

\begin{figure}
    \centering
    \includegraphics[width=\columnwidth, trim={0cm 0cm 0cm 0cm},clip]{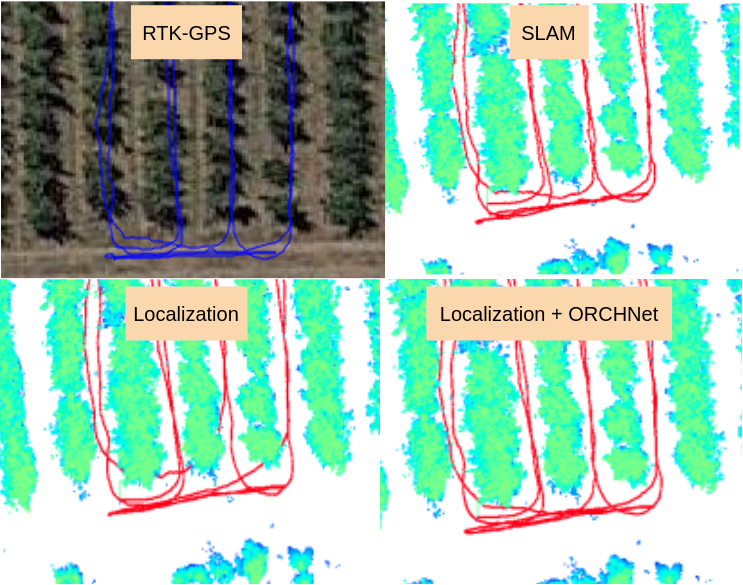}
    \caption{Highlight of qualitative evaluation results for the autumn sequence. From top to bottom and left to right: RTK-GPS data, HMaps-based SLAM, classical AMCL and ORCHNet-based AMCL.}
    \label{fig:mapsroi}
\end{figure}

To complement the results, a qualitative assessment of the proposed AMCL-ORCHNet framework is presented in Fig. \ref{fig:maps}, where it is compared with a AMCL implementation (\ie, without loop closure detection), the HMaps-based SLAM and the ground truth path obtained with an RTK-GPS.
Although the autumn sequence is a spatially small orchard, where each scan is highly overlapped, the sparse nature of the point clouds, which can be seen in Fig. \ref{fig:localization}, imposes some challenges in the registration of the scans in the map. Despite the challenges, the HMaps-based SLAM approach produces a good map representation. 
Figure \ref{fig:maps} shows that the baseline AMCL achieved, overall, a reasonable localization performance, however on the rightmost part of the geometric path (highlighted in red and depicted in Fig.  \ref{fig:mapsroi}),  the AMCL  goes through the trees in several ocations, which is corrected in the AMCL-ORCHNet approach. On the other hand, the AMCL-ORCHNet approach computes a similar path to the one obtained with the HMaps-based SLAM and the RTK-GPS in that region. These results highlight the applicability and suitability of the proposed ORCHNet as part of a localization framework in real-world conditions.

\section{CONCLUSIONS}
In this work, a 3D LiDAR-based place recognition approach (called ORCHNet) is proposed by combining the descriptors of multiple global feature aggregation methods to obtain robust global descriptors. The proposed ORCHNet is evaluated in place recognition and in an AMCL-based localization framework as a loop closure detector.

The experimental evaluation of the ORCHNet was carried out with real-world data collected in orchards during the summer and autumn seasons, using a mobile robot platform equipped with a 32-channel LiDAR. Two different task domain experiments were conducted, the first was focused on 3D LiDAR-based place recognition,  while the second explored the integration of ORCHNet in a localization framework.
Regarding LiDAR-based place recognition, ORCHNet's robustness is compared with other state-of-the-art methods and demonstrated on challenging orchard datasets, where the proposed approach demonstrates to be reliable across seasons. 
As part of a localization framework, the AMCL-ORCHNet approach corrected the path, which the AMCL baseline was not able to predict correctly.

\section*{ACKNOWLEDGMENTS}
This work has been supported by the project GreenBotics (ref. PTDC/EEI-ROB/2459/2021), founded by Fundação para a Ciência e a Tecnologia (FCT), Portugal. It was also partially supported by FCT through grant UIDB/00048/2020 and under the PhD grant with reference 2021.06492.BD. The authors would also like to thank Dr Charles Whitfield at NIAB East Malling for facilitating orchard data collection campaigns. 


\bibliographystyle{IEEEtran}

\bibliography{ref}

\end{document}